%% file: root.tex
\documentclass[letterpaper, 10 pt, conference]{ieeeconf}  %

\IEEEoverridecommandlockouts                              %

\usepackage[dvipsnames]{xcolor}

\usepackage{graphics} %
\usepackage{epsfig} %
\usepackage{mathptmx} %
\usepackage{times} %
\usepackage{amsmath} %
\usepackage{amssymb}  %
\usepackage{multirow}
\usepackage{url}
\usepackage{hyperref}

\usepackage{pifont}%
\newcommand{\cmark}{\ding{51}}%
\newcommand{\xmark}{\ding{55}}%

\usepackage{rotating}
\usepackage{adjustbox}
\usepackage{array}
\newcolumntype{R}[2]{%
    >{\adjustbox{angle=#1,lap=\width-(#2)}\bgroup}%
    l%
    <{\egroup}%
}
\newcommand*\rot{\multicolumn{1}{R{45}{1em}}}%
\usepackage{wrapfig}
\usepackage{pgfplots}
\pgfplotsset{width=7cm,compat=1.9}
\usepgfplotslibrary{external} %
\usepackage{amsmath}
\renewcommand{\vec}[1]{\mathbf{#1}}

\newcommand{\myparagraph}[1]{\par\vspace{0.3em}\noindent\textbf{#1}\hspace{0.2em}}
\newcommand{\myemphparagraph}[1]{\par\vspace{0.3em}\noindent\emph{#1}\hspace{0.2em}}

\definecolor{somegray}{rgb}{0.5, 0.5, 0.5}
\newcommand{\darkgrayed}[1]{\textcolor{somegray}{#1}}
\makeatletter
\newcommand*\titleheader[1]{\gdef\@titleheader{#1}}
\AtBeginDocument{%
  \let\st@red@title\@title
  \def\@title{%
    \vskip-3em
    \bgroup\normalfont\large\centering\@titleheader\par\egroup
    \vskip1em\st@red@title}
}
\makeatother

\titleheader{\darkgrayed{This paper has been accepted for publication at the 
IEEE/RSJ International\\Conference on Intelligent Robots and Systems (IROS), Abu Dhabi, 2024.
\copyright IEEE}}

\title{\LARGE \bf
Deep Visual Odometry with Events and Frames
}

\author{Roberto Pellerito,$^{1}$  Marco Cannici,$^{1}$ Daniel Gehrig,$^{1}$ Joris Belhadj,$^{2}$ Olivier Dubois-Matra,$^{2}$ \\ Massimo Casasco,$^{2}$ Davide Scaramuzza$^{1}$\\ \\
$^{1}$ Robotics and Perception Group, University of Zurich, Switzerland\\
$^{2}$ European Space Agency
\thanks{This work was supported by the European Union’s Horizon Europe Research and Innovation Programme under grant agreement No. 101120732 (AUTOASSESS) and the European Research Council (ERC) under grant agreement No. 864042 (AGILEFLIGHT).}
}

\begin{document}

\maketitle
\thispagestyle{empty}
\pagestyle{empty}

\input{sections/0.abstract}
\input{sections/1.introduction}

\input{sections/2.related_works}
\input{sections/3.method}
\input{sections/4.experiments}
\input{sections/5.conclusions}

\bibliographystyle{IEEEtran}
\bibliography{IEEEabrv,bibliography,all}

\end{document}

%% file: sections/0.abstract.tex
\begin{abstract}
Visual Odometry (VO) is crucial for autonomous robotic navigation, especially in GPS-denied environments like planetary terrains. 
To improve robustness, recent model-based VO systems have begun combining standard and event-based cameras. While event cameras excel in low-light and high-speed motion, standard cameras provide dense and easier-to-track features.
However, the field of image- and event-based VO still predominantly relies on model-based methods and is yet to fully integrate recent image-only advancements leveraging end-to-end learning-based architectures. Seamlessly integrating the two modalities remains challenging due to their different nature, one asynchronous, the other not, limiting the potential for a more effective image- and event-based VO.
We introduce RAMP-VO, the first end-to-end learned image- and event-based VO system. It leverages novel Recurrent, Asynchronous, and Massively Parallel (RAMP) encoders capable of fusing asynchronous events with image data, providing $8\times$ faster inference and $33\%$ more accurate predictions than existing solutions.
Despite being trained only in simulation, RAMP-VO outperforms previous methods on the newly introduced Apollo and Malapert datasets, and on existing benchmarks, where it improves image- and event-based methods by $58.8\%$ and $30.6\%$, paving the way for robust and asynchronous VO in space.
\end{abstract}

%% file: sections/1.introduction.tex
\vspace{12pt}
\noindent\textbf{Multimedial Material:} For code and datasets visit \url{https://github.com/uzh-rpg/rampvo}.

\section{Introduction}
\label{sec:intro}

Visual Odometry (VO) is vital for robotic platforms but often fails in challenging scenarios involving low-light, high dynamic range, or high-speed motion. These shortcomings are often caused by the limitations of traditional frame-based camera, such as their susceptibility to motion blur, limited dynamic range, and unfavorable bandwidth-vs-latency tradeoff. While higher frame rates reduce latency, they come at the cost of higher bandwidth and increased processing power. Event-based cameras, which record per-pixel brightness changes asynchronously, address all these issues, offering high dynamic range (HDR), low latency, and low bandwidth and power usage, making them the ideal complement to regular cameras in VO systems.
Their combination holds significant promise for critical VO applications, especially when sensors like GPS, LiDAR, and Inertial Measurement Units (IMUs) cannot be used or are ineffective due to radiation and temperature changes. These conditions are often encountered in planetary exploration and landing missions, where rapid motion and partial shadow are also common.

\begin{figure}[t]
    \centering
    \includegraphics[width=\linewidth]{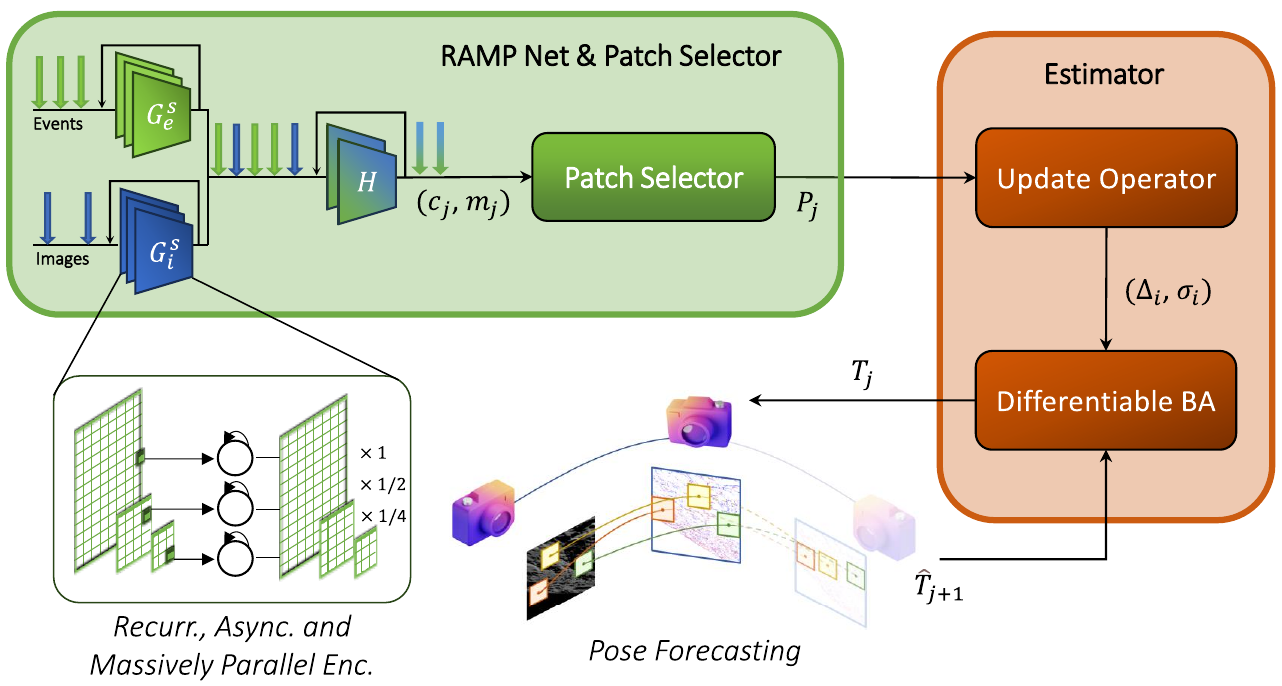}
        \caption{Recurrent, Asynchronous, and Massively Parallel (RAMP) Encoders are used to process asynchronous events and images.
    Patches are extracted from the resulting encoding and used by the Estimator inspired by DPVO~\cite{dpvo} to perform data-driven feature tracking and visual odometry. A simple pose forecasting module exploits previously extracted patches to initialize poses in the bundle adjustment, allowing for improved performance.}
    \label{fig:Overview}
    
\end{figure}

Recent model-based solutions \cite{eds} have shown potential in fusing images and events. However, the fields of image-only \cite{dpvo,droidslam} and event-only \cite{klenk2023deep} VO have recently shown that learning-based pipelines, trained end-to-end, can surpass traditional model-based systems in accuracy and robustness.
While the combination of data-driven approaches with systems that leverage images and events appears promising, effectively combining event data---with its distinctive asynchronicity and sparsity---with synchronous and dense frames is a non-trivial challenge in learning-based solutions. Traditional learning-based methods typically resort to artificially synchronizing events at image timestamps to facilitate data fusion, reducing the rate at which events are processed to that of the slower image modality.
Nevertheless, in tasks such as VO, this simplification is not ideal and might limit the algorithm's ability to exploit events received in between images, which is crucial for tracking features effectively. 

To address these limitations, our work introduces an adaptive fusion approach that adjusts the frequency of event fusion based on the rate of incoming events, thus mirroring the pace of the scene dynamics. Our Recurrent, Asynchronous, and Massively Parallel (RAMP) encoders handle asynchronous events and images at varying rates and fuse them into a pyramidal memory that serves as a data-agnostic feature space. We use these encoders in RAMP-VO, the first learning-based VO method that uses both events and frames. RAMP-VO leverages a motion-aware strategy based on event data to extract robust patch-based feature tracks. These features are then processed by a differentiable bundle adjustment module \cite{dpvo} which leverages a simple pose forecasting module for initialization.

We train RAMP-VO on an event-based version of TartanAir~\cite{tartan_air}. To address the lack of visual odometry datasets that feature image and event data in challenging space landing settings, we also introduce two novel datasets: the Malapert landing and the Apollo landing datasets, which feature challenging motion and lighting conditions due to stark shadows cast by the sun.
The first dataset represents a realistic simulation of a spacecraft landing, covering several kilometers of descent near the Malapert crater in the south Moon pole. %
The second dataset, captured with real RGB and event cameras, features landings on a 3D scale model of the lunar surface and precise ground truth camera poses, making it a valuable resource for research and evaluation.

Despite being trained purely in simulation, RAMP-VO outperforms both image-based and event-based methods on traditional real-world benchmarks, as well as on the newly introduced Apollo and Malapert landing datasets.
To summarize our contributions are:

\begin{itemize}
    \item A novel massively parallel feature extractor, termed RAMP encoder, that asynchronously\footnote{Notice that in this work, the term “asynchronous” refers to our network's ability to handle data streams at different and varying rates. Our network processes images or packets of events (as frame-like event representations) as soon as they are available, without stream synchronization. This differs from networks operating on event-by-event processing, where “asynchronous” refers to the network layers' functioning.} fuses images and events, both spatially and temporally. Our encoder is 8 times faster and achieves a 33\% higher performance than other state-of-the-art asynchronous solutions. 
    \item RAMP-VO, the first learning-based VO using events and frames, which outperforms both image-based and event-based methods by $58.8\%$ and $30.6\%$, respectively on traditional real-world benchmarks.
	\item Two novel datasets, Apollo and Malapert landing, targeting challenging planetary landing scenarios.
\end{itemize}

%% file: sections/2.related_works.tex
\section{Related Work}
\label{sec:related_works}

\textbf{Learning-based Visual Odometry. } Recent advancements in VO have witnessed a paradigm shift towards learning-based approaches \cite{chen2023deep}, surpassing traditional methods in accuracy and robustness \cite{song2023contextavo,kazerouni2022survey}. Unsupervised methods \cite{ranjan2019competitive,lu2023deep,zhao2023self} exploit additional depth and optical flow predictors while recent methods employ neural radiance fields \cite{zhu2022nice,sucar2021imap,sandstrom2023uncle} to further improve performance. Supervised methods, on the other hand, either rely on end-to-end camera-motion regressors \cite{wang2017deepvo,xue2019beyond,saputra2019distilling,kuo2020dynamic} or exploit hybrid solutions that combine geometric models with deep neural networks \cite{sun2022improving,zhan2020visual,yang2020d3vo}. Among these works, DROID-SLAM \cite{droidslam} recently proposed to combine an iterative learning-based optimization inspired by RAFT \cite{raft} with a differentiable bundle adjustment layer. The follow-up work, DPVO \cite{dpvo}, further improves its efficiency by replacing dense feature tracking with a sparse patch-based variant. Our work builds upon DPVO, but significantly improves its robustness by effectively fusing images and events together.

\textbf{Event-based Motion Estimation. } Although full 6DOF pose estimation using only events has been successfully demonstrated in the literature \cite{kim2016real,Rebecq2017EVOAG}, most event-based VO rely on additional sensors. While some systems incorporate depth estimates from stereo or depth cameras \cite{esvo,zuo2022devo, Event_based_3D_SLAM}, others integrate IMU measurements to improve robustness and scale recovery \cite{ultimateslam,zihao2017event,Rebecq17bmvc,eklt-vio,Guan2022PL_EVIO,chen2023esvio,zihao2017event}. 
Standard image frames have also been incorporated to extract features and then track them with events \cite{EventbasedFeatureTracks} or to optimize additional residual errors \cite{ultimateslam,eds}, often exploiting DAVIS cameras \cite{Brandli14ssc} or beamsplitter setups~\cite{eds}.

Despite these advancements, challenges persist in low-texture environments, directing toward exploration into deep learning approaches. While early attempts focused on unsupervised techniques \cite{zhu2018unsupervised,ye2018unsupervised}, DEVO \cite{klenk2023deep} has recently shown promise in transferring to out-of-distribution scenarios by training on large simulated datasets. Notably, however, DEVO only utilizes events and still depends on encoders primarily optimized for images. In contrast, our work employs a recurrent and pyramidal feature extractor that effectively fuses images with events, preserving their incremental nature and exploiting the best of both modalities.

\textbf{Fusing events and frames. } While a great variety of approaches leverage, optimize, or fuse images and events for different downstream tasks \cite{Alonso19cvprw,Tulyakov2021cvpr,Tulyakov22cvpr,Sun2022cvpr},  the topic of effectively fusing the two data modalities while considering their different nature has thus far been underexplored.
Some methods \cite{Alonso19cvprw,Hu2020ITSC,Tulyakov2021cvpr} synchronize and concatenate both modalities and process them together with a shared encoder. Others \cite{Tulyakov22cvpr,messikommer2023datadriven}, instead, use specialized feature extractors, but still resort to data synchronization for processing. 

To date, only one prior study, RAM Net \cite{ramnet}, has proposed a specialized and asynchronous way of fusing events and frames. However, RAM Net relies on a sequential hierarchical feature extraction process and 
utilizes slow Conv-GRU modules. %
Our asynchronous encoders build upon RAM Net but exploit pixel-wise operations and parallel extraction of multi-scale features, demonstrating both higher performance as well as improved efficiency. %

%% file: sections/3.method.tex
\section{Methodology}
\label{sec:method}

\begin{figure*}
	\center
	\includegraphics[width=\textwidth]{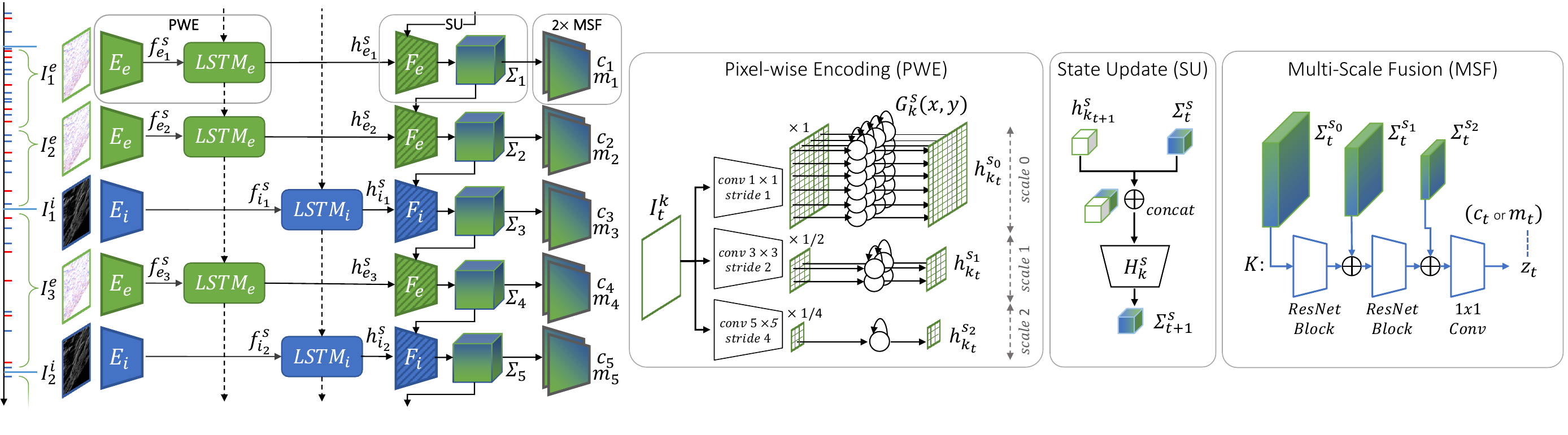}
 \vspace{-5ex}
	\caption{An overview of the proposed RAMP Net encoder. Events and images are first asynchronously processed by two parallel pixel-wise, multi-scale, encoding branches (PWE) made of a set of convolutional layers followed by pixel-wise LSTMs $G_k^s$. A shared state  $\Sigma^s_t$ is then updated (SU) with features coming from different data modalities $k$ by employing sensor-specific encoders at each scale. The multi-scale features are then finally combined through two separate fusion modules (MSF) to produce the matching and context features, $m_t$ and $c_t$}
	\label{fig:Multiscale_encoder}
\end{figure*}

Our end-to-end event- and frame-based visual odometry algorithm, RAMP-VO, builds on deep patch visual odometry (DPVO)~\cite{dpvo} and takes inspiration from recurrent asynchronous multimodal (RAM) networks~\cite{ramnet}. The main innovations of this work reside in how events and frames are fused together. An ideal vision encoder should fuse sensor measurements at an adaptable rate, processing more information during fast motion for better tracking, and ensuring reliability even when a sensor experiences temporary outages. Our RAMP encoders achieve this by creating a sensor-agnostic representation that is recurrently updated whenever new data becomes available, either images or events. We present the main components of RAMP-VO in Section \ref{sec:method_overview}, and the RAMP encoder in Section \ref{sec:encoder}.

\subsection{Overview} \label{sec:method_overview}
RAMP-VO processes a temporally ordered, asynchronous stream of data made of `frames'. These are either standard images or frames made of events with $H \times W$ resolution. Similar to DPVO~\cite{dpvo}, we extract a set of patches and predict their optical flow across frames by computing correlation volumes over \emph{matching features} $\vec{m}_j$ and additional \emph{context features} $\vec{c}_j$. We iteratively refine the predicted patch flow and finally use a differentiable bundle adjustment layer to update the camera poses. For each new frame with index $j$ we first compute $\vec{m}_j$ and $\vec{c}_j$ through the RAMP encoder described in Section \ref{sec:encoder}, and then extract $N$ patches with dimension $p\times p$ from these feature maps.
While DPVO \cite{dpvo} and DEVO \cite{klenk2023deep} select these patches randomly or through a learning-based strategy, we opt for a simpler yet effective alternative. In particular, we rely on events' activity to extract patches from textured areas, as these regions usually generate a higher number of events compared to uniform ones. 
We first compute an $H \times W$ density map by counting the number of events per pixel. We then apply non-maximum suppression to remove candidates that are not the highest in their $11\times11$ neighborhood and finally obtain the patch centers by selecting the $N$ locations with the highest counts.

We denote the $l$-th patch $\vec{P}^l_j$ extracted from frame $j$ as 
\begin{equation}
    \vec{P}^l_j = \begin{bmatrix}
        \vec{x}&
        \vec{y}&
        \vec{1}&
        \vec{d}
    \end{bmatrix}^T,  \quad \mathbf{x},\mathbf{y},\mathbf{d}\in \mathbb{R}^{1\times p^2}.
\end{equation}
As in DPVO, we model patches as collections of contiguous pixels, with $\textbf{x}, \textbf{y}$ the coordinates of the pixels in the patch and $\textbf{d}$ their depth (constant within the patch).

\begin{figure}
	\center
	\includegraphics[width=0.9\linewidth]{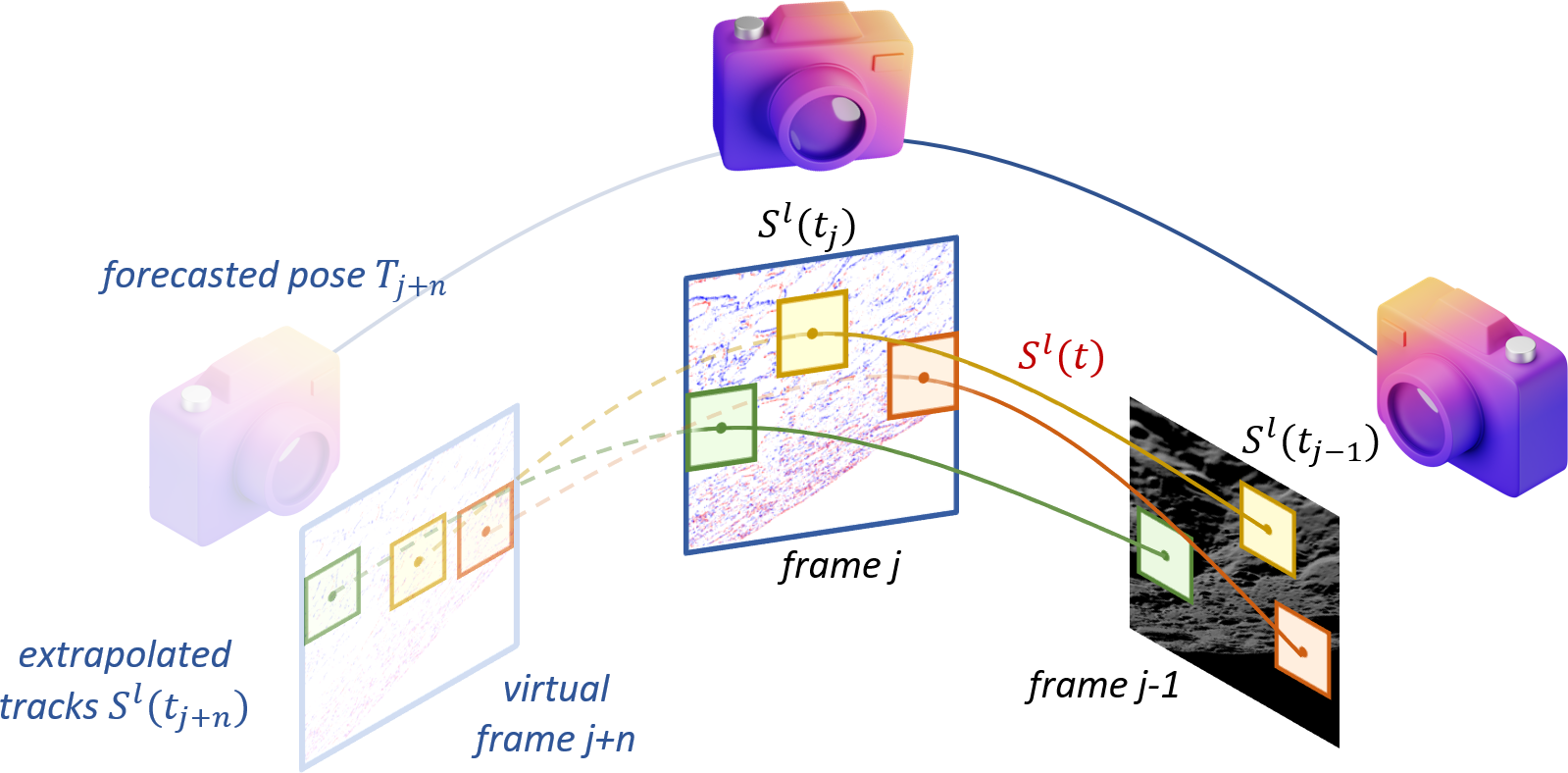}
	\caption{Illustration of pose initialization. Through patch extraction and projection into future frames we construct feature tracks for frames $j,j-1,...$ which we use to construct the splines $S^l(t_j)$. To perform pose initialization, we extrapolate the feature tracks to time $t_{j+n}$, and apply bundle adjustment to solve for the forecasted pose $T_{j+n}$.}
	\label{fig:poseForecasting}
\end{figure}

Following \cite{dpvo}, we build a bipartite \emph{patch graph} $\mathcal{E}$,  depicted in Figure \ref{fig:poseForecasting}, by connecting a patch $\vec{P}^l_j$ extracted from frame $j$ to every frame $i$ within a distance $r$ from $j$, each containing the projection $\vec{P'}^l_{ji}$ of the original patch onto the frame:
\begin{equation}
    \vec{P'}^l_{ji} \sim \vec{\bar{K}}\vec{T}_i\vec{T}_j^{-1}\vec{\bar{K}}^{-1}\vec{P}^l_j, \quad
    \vec{\bar{K}} = \begin{pmatrix} \vec{K} & \vec{0} \\ \vec{0}^T & 1 \end{pmatrix}
\end{equation}
where $\vec{\bar{K}}$ is a $4\times 4$ matrix built from the $3\times3$ camera matrix $\vec{K}$ and $\vec{T}_{i},\vec{T}_{j}$ are poses of frames $i,j$. We summarize this operation as $\vec{P'}^l_{ji} = \omega(\vec{T}_i,\vec{T}_j, \vec{P}^l_j)$.

Next, RAMP-VO computes camera motion by estimating 2D corrections $\boldsymbol{\Delta}_{li}\in\mathbb{R}^2$ for each projected patch $\mathbf{P'}^l_{ji}$, as well as importance values $\sigma_{li}\in\mathbb{R}^{2\times 2}$ through a series of blocks involving a correlation lookup, 1D Convolution, Soft Aggregation, Transition Block and Factor Head. These steps extract the features of each patch in frame $j$ from $\vec{m}_j$ and compare them with those obtained by cropping $\vec{m}_i$ with the reprojection of the patch in frame $i$, together with additional context features $\vec{c}_i$. Since these operations, described in \cite{dpvo}, are out of the scope of the current work, we summarize them with the \emph{update operator} $F$ in Fig. \ref{fig:Overview}:
\begin{equation}
    \Delta_{li}, \sigma_{li} = F(\vec{P'}^l_{ji}, \vec{c}_i, \vec{m}_i, \vec{m}_j)
\end{equation}

Finally, given the corrected positions ${\mathbf{P'}}^l_{ji} + \Delta_{li} $ and their weights $\sigma_{li}$, RAMP-VO performs a differentiable bundle adjustment (BA) step, which minimizes the projection error:  
\begin{equation}
\label{eq:ba}
    \sum_{(l,i)\in\mathcal{E}} \left\Vert \left[ \hat{\mathbf{P'}}^l_{ji} + \Delta_{li} \right] - \hat{\omega}(\mathbf{T}_i,\mathbf{T}_j, \mathbf{P}^l_j)\right\Vert^2_{\sigma_{li}}.
\end{equation}
Here the left-hand side is kept fixed, while the optimization solves for the camera poses $\mathbf{T}_i, \mathbf{T}_j$ and depths $\mathbf{d}^l_j$, refining the camera trajectory to match the predicted patch trajectories. The $\hat{}$ operator selects the central pixel of the patch, and $\mathcal{E}$ is the path graph. %
This operation is fully differentiable, and thus used during training to backpropagate errors. 

Contrary to \cite{dpvo} which uses a simple linear initialization, we bootstrap initial poses in our RAMP-VO by fitting two cubic univariate splines modeling the 2D motion of the patch center along the last $11$ frames, such that $( \mathbf{x}^l_j, \mathbf{y}^l_j ) \sim \mathbf{S}^l(t_j) = ( S^l_x(t_j), S^l_y(t_j) )$. We then extrapolate the location of the camera at time $t_{j+1}$, by evaluating $\mathbf{S}^l(t_{j+1})$, assuming the first and second derivatives to be zero at the splines' boundaries and constant depth. We obtain the 6 DOF pose at time $t_{j+1}$ through BA, optimizing Equation \ref{eq:ba} in which we use $\mathbf{S}^l(t_{j+1})$ in place of the corrected patch position.

\subsection{Asynchronous and Massively Parallel Encoders}
\label{sec:encoder}
Denote the stream of data $\{\vec{I}_{k_j}(t_j)\}_{j=1}^T$ captured at timestamps $t_j$. 
Here $\vec{I}$ (henceforth denoted as "frame") denotes either a $5\times H\times W$ sized event stack ~\cite{Mostafavi19cvpr}, in the case of events, or a $C\times H\times W$ sized image\footnote{For color images $C=3$ and for gray-scale images $C=1$.}. The variable $k_j \in \{\text{e}, \text{i}\}$ denotes the sensor at timestamp $t_j$. 
We encode these data structures using a Recurrent, Asynchronous and Massively Parallel (RAMP) encoder. An overview of the architecture is provided in Figure \ref{fig:Multiscale_encoder}. We employ two different Multi-Scale Fusion modules, with identical structures, to generate either the context or the matching feature maps.

We designed the RAMP encoders to be easily parallelizable and to effectively fuse asynchronous data streams. To do so, we encode each pixel independently, in contrast with ~\cite{Gehrig21ral} which requires sequential processing and expensive two-stage ConvGRU operations, and recurrently fuse information into a shared state. First, RAMP encoders process the data stream with pixel-wise, sensor-specific networks at multiple scales $s$, creating a sequence of features $\{\vec{f}^s_{k_j}\}_{j=1}^T$:
\begin{equation}
    \vec{f}^s_{k_j}(t_j) = E^s_{k_j}(\vec{I}_{k_j}(t_j)).
\end{equation}
These encoders have a kernel size $1\times1$, $3\times3$, and $5\times5$ respectively, and a stride of $2^s$ with $s=0,1,2$. %

This stream of features is then further encoded into a sensor-agnostic state through two recurrent stages: (1) intra-sensor update, and (2) inter-sensor update. 
In the intra-sensor update step, we process features originating from a single sensor and scale using an LSTM operating on individual pixels, inspired by \cite{cannici2020differentiable}
\begin{equation}
\vec{h}^s_{k_j}(t_j) = G^s_{k_j}(\vec{f}^s_{k_j}(t_j)),
\end{equation}
where we have omitted the cell state for brevity.

The inter-sensor update step, finally combines together the hidden states from separate sensors asynchronously using only a single depth-wise convolution in the following way:
\begin{equation}
    \boldsymbol{\Sigma}^s_j = H_{k_j}\left([\vec{h}^s_{k_j}(t_j) \Vert \Sigma_{j-1}]\right)
\end{equation}
where $\Vert$ denotes concatenation. A different depth-wise convolution $H_{k_j}$ is used depending on which sensor the data being encoded is generated from. Notably, updates to the $\boldsymbol{\Sigma}^s_j$ state can occur at any frequency, enabling our encoder to seamlessly handle asynchronous data streams.

After generating the sensor-agnostic states $\Sigma_j$ at each time $t_j$, two hierarchical encoders generate either the context or the matching feature maps, each at $1/4$ the original resolution
\begin{equation}
    \vec{z}_j = K(\{\Sigma_j^s\}_{s=0}^2).
\end{equation}
Here $\vec{z}_j$ can be either $\vec{c}_j$ or $\vec{m}_j$, \emph{i.e.}, the context and matching features required by the DPVO backend.

\myparagraph{Multi-Scale Fusion. }
The Multi-scale Fusion (MSF) module follows the single-scale feature extractors in \cite{dpvo} and consists of a $7\times7$ convolution with stride $2$, four residual blocks (two at $1/2$ and two at $1/4$ of the initial resolution), and a final $1\times1$ convolution. We follow the configuration in \cite{dpvo}, using full-scale features $\Sigma_j^0$ as input to the encoder, but additionally injecting $\Sigma_j^1$ and $\Sigma_j^2$ after the second and fourth residual blocks respectively. We do so by concatenating these features to the residual block's outputs and adjusting the channels of the next operation to accommodate the additional features.

%% file: sections/4.experiments.tex
\section{Experiments}
\label{sec:experiments}

\begin{figure*}
  \centering
  \begin{tabular}{ccc}
    \includegraphics[width=0.3\textwidth]{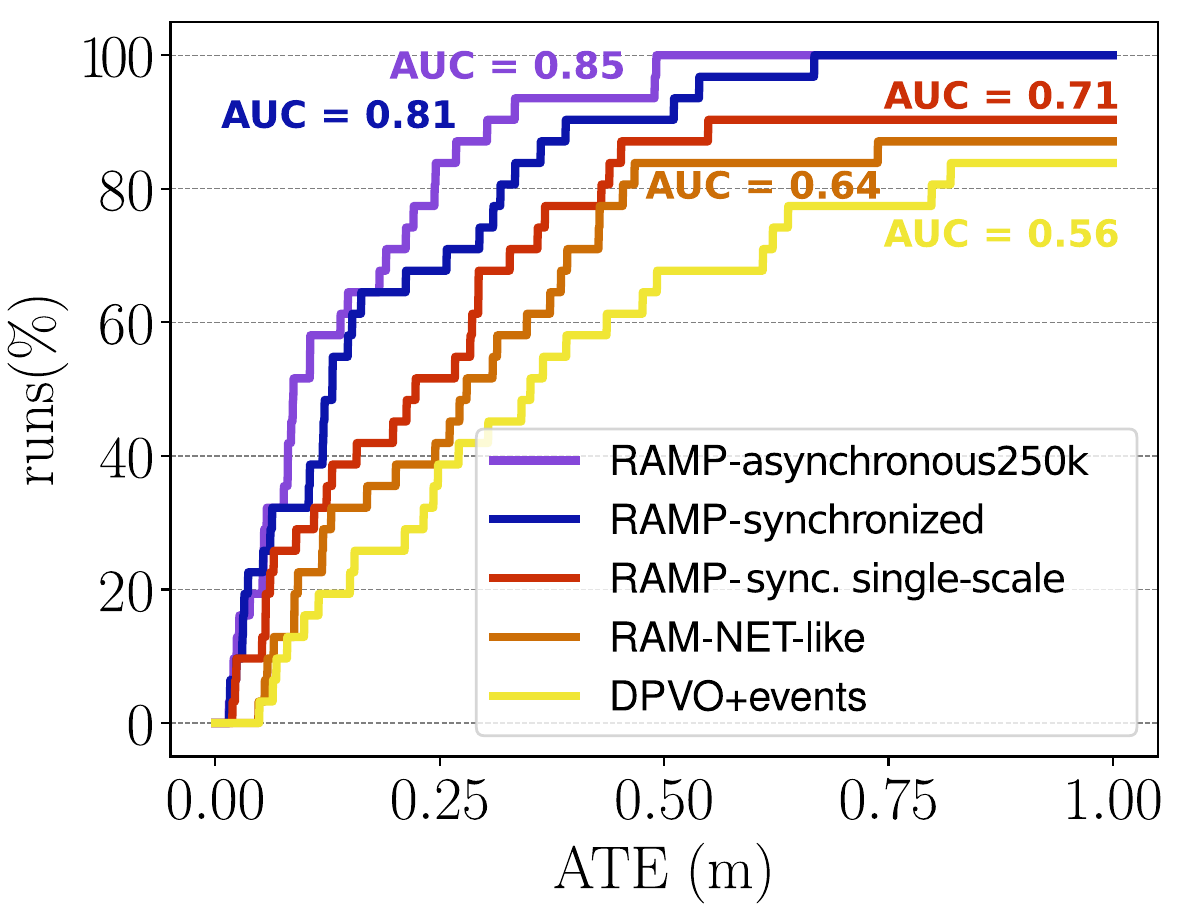}\vspace{-1pt}&
    \includegraphics[width=0.3\textwidth]{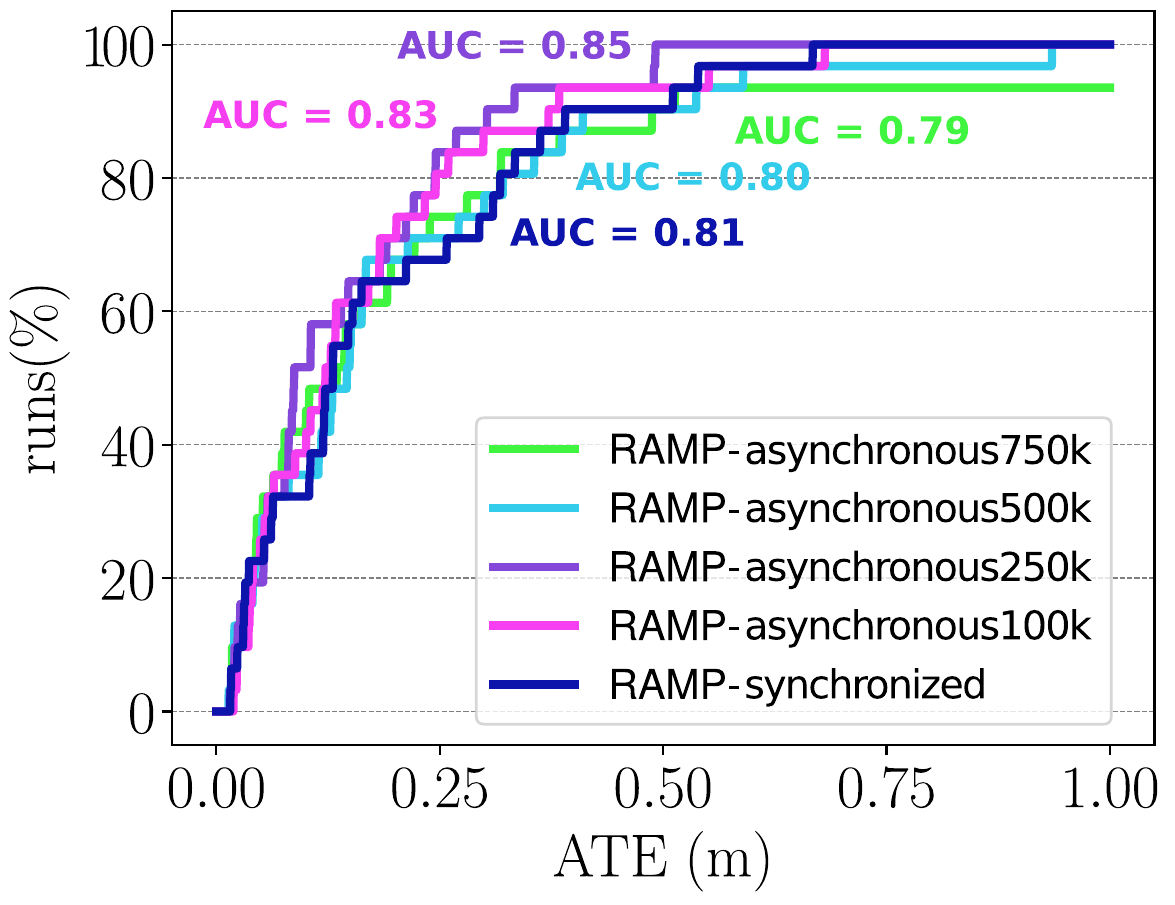}\vspace{-1pt}&
    \includegraphics[width=0.3\textwidth]{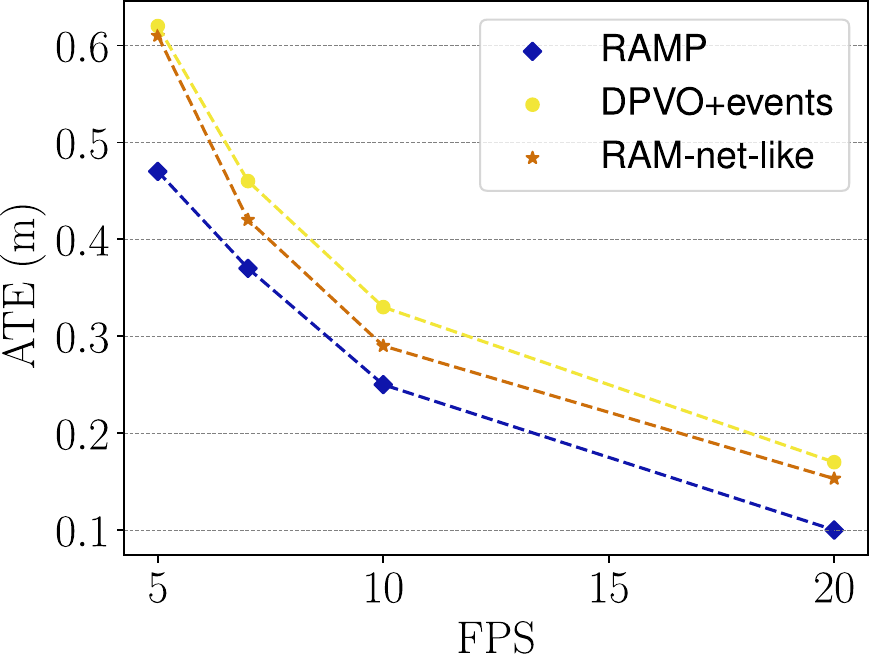}\vspace{-1pt}\\
    {\small (a) Ablation Study} & {\small (b) Asynchronous Processing} & {\small (c) Low Framerate VO}
  \end{tabular}

    \caption{
    Comparisons of the ablated models (a) and of RAMP with asynchronous and synchronized data (b) on the full TartanAir test set.  We show the importance of using a RAMP encoder as in RAMP-VO over a sequential, single-scale encoder and feed-forward encoder. We also show the benefits of using the full event information in RAMP-VO with finer discretizations. The RAMP encoder is better at maintaining memory than the RAM-Net-like encoder, as highlighted in low framerate VO experiments (c) on the \emph{carwelding} sequences of TartanAir.}
    
  \label{fig:roc}
\end{figure*}

\textbf{Training.}
We train RAMP-VO on the synthetic TartanAir \cite{tartan_air} dataset, with same train and test-split in \cite{dpvo}, which we augment with events using VID2E \cite{Gehrig20cvpr}. 
During training, we follow the DPVO \cite{dpvo} training scheme to select and filter frames for training.
Additionally, we also enforce a minimum of $1.2$M events between every pair of frames and remove additional frames if this condition is not satisfied.
Finally, we feed RAMP-VO by interleaving two event stacks for every pair of selected frames. We create the first event stack by stacking the $600,000$ events received just before the mid-timestamp between the pair of images, and the second event stack by aggregating the $600,000$ events preceding the second frame. Since ground truth poses are only given for frames, we do not compute a loss for the mid-frame events. 
At inference time, we feed the model with asynchronous images and events, by creating a new event stack each time $M$ events are received. As the number of triggered events depends on the scene dynamics, with more events triggered with faster motion, the number of event stacks collected varies adaptively. For comparisons, we also test our model using the data feeding strategy used for training, which we call \emph{synchronized} as events are collected at regular rates.

We train RAMP-VO for $350,000$ steps with sequences of $15$ images and $30$ event stacks on a Quadro RTX 8000 GPU. The remaining hyperparameters are set as in DPVO \cite{dpvo}. Full training takes around $8$ days on our hardware. 

\begin{figure*}
  \centering
  \begin{tabular}{cc}
  \includegraphics[height=2.8cm]{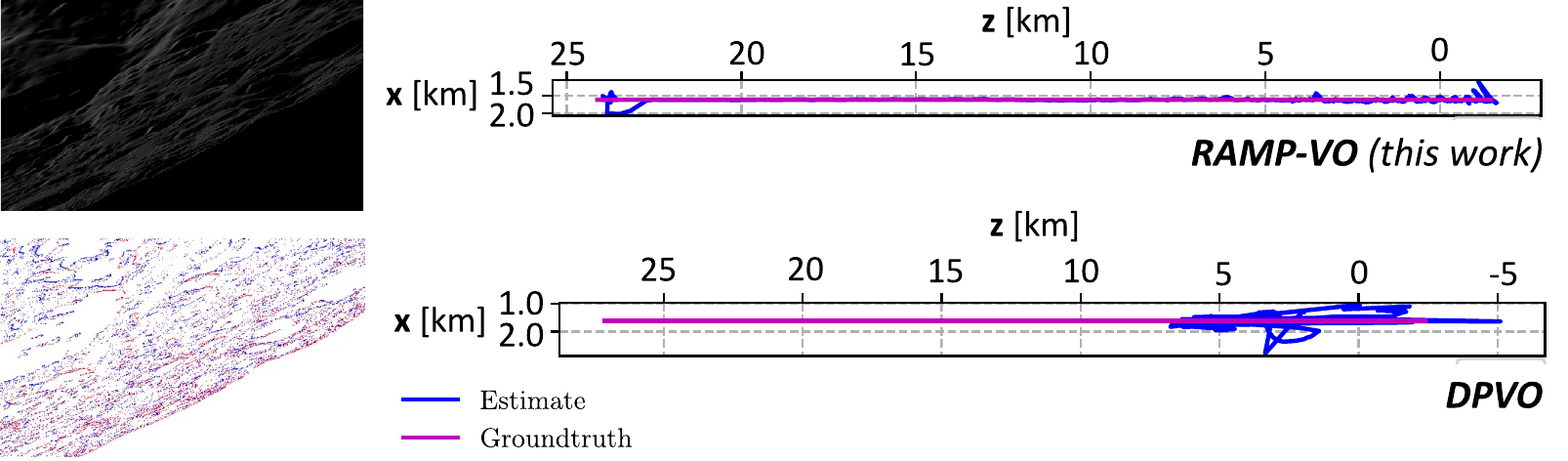}&
  \includegraphics[height=2.8cm]{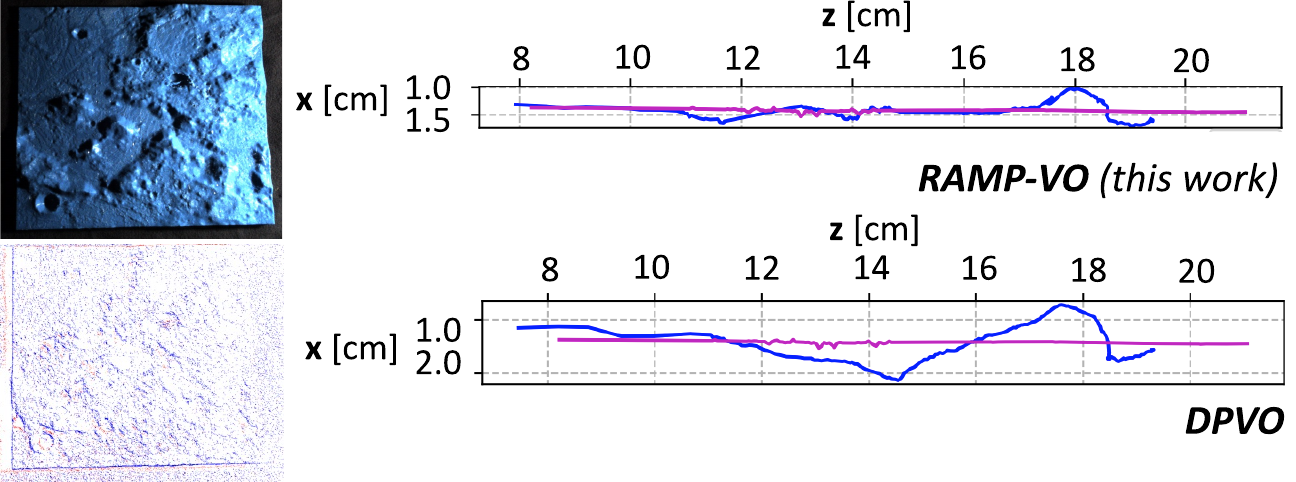}\\
  (a) Malapert dataset& (b) Apollo dataset
  \end{tabular}
  \caption{Preview, and qualitative trajectory comparison on Malapert dataset (a), and Apollo dataset (b). Note that while the Malapert sequence is measured in kilometers, the Apollo sequence, recorded at a miniature scale of the Moon's surface, is in centimeters.}
  \label{fig:qualitative_malapert_apollo}  
\end{figure*}

\myparagraph{Datasets.}
We use the Stereo DAVIS \cite{Zhou18eccv} and EDS \cite{eds} datasets, and the ECCV 2020 SLAM competition TartanAir \cite{tartan_air} test-split to benchmark our method. Stereo DAVIS and EDS are real-camera datasets, allowing us to test RAMP-VO's robustness to domain shift and noise originating from the sensor or potential miscalibration.
To ablate the crucial components of RAMP-VO architecture, we use instead the TartanAir test-split used in \cite{dpvo}.
Additionally, we introduce two new benchmarks that replicate lunar surface landings and feature challenging lighting conditions: Apollo landing and Malapert landing. These benchmarks feature rapid motion, high dynamic range, and textureless terrains.

\myemphparagraph{Malapert landing.} It features a total of $20$ minutes of simulated recordings from the Malapert south Moon region, divided into $2$ sequences.
We use the planets and satellites simulator PANGU \footnote{https://pangu.software/}\label{pangu} to generate realistic descent trajectories, each $250$ km long in altitude and $40$ km in translation, featuring partial or complete darkness.
Ground truth poses of the spacecraft's center of mass are provided at $5$Hz, together with $640 \times 480$ synchronized RGB images. We generate synthetic events using Vid2E~\cite{Gehrig20cvpr} with default settings.

\myemphparagraph{Apollo landing.}
A real dataset consisting of a total of $5$ minutes of recording, split into $6$ trajectories, featuring both vertical and lateral descent trajectories on a $260 \times 260$ cm scale replica of the Apollo 17 landing site.
Frames and events are recorded with a beam-splitter 
similar to the one used in \cite{eds}. Poses are recorded with an OptiTrack motion capture system. We downsample frames and events to a resolution of $640\times 480$ before processing.

\myparagraph{Baselines.}
We evaluate the proposed RAMP-VO architecture against several VO state-of-the-art methods making use of images only (I), events only (E) as well as based on both images and events (I+E).
 We follow the evaluation in \cite{eds, dpvo, klenk2023deep} and select ORB-SLAM2 \cite{MurArtal17tro}, ORB-SLAM3 \cite{orbslam3}, COLMAP \cite{Schonberger_2016_CVPR, schoenberger2016mvs}, DROID \cite{droidslam} and DPVO \cite{dpvo} as image-only baselines, while we use EDS \cite{eds} for comparison against methods fusing images and events, being the only VO system of this kind in the literature. 
We implement an event-only DPVO baseline, EDPVO, that directly processes event stacks, as well as one that processes images and events concatenated together, DPVO+events. 
Finally, we also compare against DEVO \cite{klenk2023deep} which makes use of a similar DPVO-based architecture but only relies on events, contrary to our method that also uses images.

\subsection{Effects of RAMP blocks}
We start by analyzing the individual contribution of RAMP-VO modules and input modalities on the TartanAir \cite{tartan_air} test set.
In this section, we adopt the evaluation protocol in~\cite{dpvo}, which analyzes the percentage of sequences from the TartanAir test set below a given absolute trajectory error threshold, producing plots like the one in Fig.~\ref{fig:roc}. This is done to discount the effect of individual diverging sequences, which report abnormally high trajectory errors, and would skew the average. 
We use 5000 different thresholds equally spaced between an ATE[m] of 0 to 1.
We summarize the results by computing the the area under the curve (AUC), and use it to compare between ablations.

\textbf{Results:}
From Figure \ref{fig:roc} (a), DPVO+events has the lowest performance among the methods we tested (AUC of $0.56$), highlighting that a trivial event-image fusion is not sufficient to exploit events. Moreover, our RAMP-VO encoder in synchronous mode outperforms the baseline that uses RAM-Net both when we use multiple scales, as in RAM-Net, but also when just one scale is used. The single-scale RAMP-VO (AUC of $0.71$) achieves a 11\% increase over the RAM-Net encoder (AUC $0.64$), while the multi-scale version (AUC $0.81$) improves by 27\% over RAM-Net. When events are processed asynchronously, using $M=250000$, the performance increases even more, reaching a $0.85$ AUC and a 33\% improvement.

In Figure \ref{fig:roc} (b), we further analyze the performance of RAMP-VO using asynchronous images and events while varying the number of events $M$ from 100000 to 750000.
Although RAMP-VO is trained with synchronized events stacks and frames, it demonstrates consistent performance with asynchronous data. Notably, as the event sequence's discretization becomes finer (i.e., smaller $M$ values), performance improves, peaking at $M=250,000$, where it achieves a $5\%$ improvement compared to synchronous event feeding.

\textbf{Low Framerate VO:}
In this section, we demonstrate that our method is not only able of asynchronous processing, but it can also cope with frames provided at very low rates.  Since our RAMP encoders build sensor-agnostic features, we can exploit the faster data stream, i.e., event voxels, to keep the feature embedding updated, and generate a consistent pose.

To test this capability, we design an experiment on the \emph{carwelding hard p003} and \emph{carwelding easy p007} sequences of the TartanAir, where we artificially reduce the framerate of the images, subsampling them, but keeping event stacks fixed at 20 Hz.  
We then evaluate the trajectory error against ground truth poses at 20 Hz for DPVO+events, RAMP-VO, and RAMP-VO with RAM-Net-like encoder, and report the results in Fig.~\ref{fig:roc} (c). 
Our RAMP encoder consistently outperforms the baselines, including the RAM-Net-like encoder, indicating its superior performance on asynchronous data. 

\textbf{Timing Results:} To further motivate the use of the proposed RAMP encoders, as opposed to RAM Net \cite{ramnet}, we time the two encoders on the TartanAir test set. We measure the average time required to extract features $\Sigma_j$ from a single frame (or event) on a Quadro RTX 8000 GPU. RAM Net takes $370$ ms on average, while the proposed RAMP Net only takes $47$ ms, resulting in a $8\times$ speedup. Leveraging pixel-wise feature processing, our encoder exploits significantly higher parallelism than RAM Net. %

\subsection{Results on Space Data}
Next, we validate RAMP-VO on the low light and low frame rates Malapert and Apollo landing datasets, where %
report the average absolute trajectory error over 5 runs. %

\input{tables/tab_malapert_apollo}
\input{tables/tab_tartanAIR_competition}
\input{tables/tab_stereo_davis}

\textbf{Results:} In Table \ref{tab:malapert_apollo}, we report performance on the Malapert landing dataset, where
RAMP-VO recovers accurate poses deviating only 0.2\% to 1.7\% from the ground truth over 250km trajectories.
DPVO, instead, can not recover a valid trajectory leading to ATE errors of several kilometers, equivalent to $20\%$ to $30\%$. When events are added, DPVO decreases the error by $45.14\%$ to $29.46\%$, highlighting the importance of events in dark regions. We report a qualitative comparison on a Malapert sample in Figure \ref{fig:qualitative_malapert_apollo}.

On the Apollo landing dataset, both RAMP-VO versions, 
multi- and single-scale (SS),
outperform image and image+event DPVO baselines by up to $30.77\%$, reaching an error from $2\%$ to $6\%$ of the ground truth.
Single-scale and multi-scale RAMP-VO achieve similar results on Apollo, contrary to Malapert's low-light environments, where single-scale errors are twice as high, indicating the need to focus on both global and fine-grained details for improved robustness.

\subsection{Comparison with State of the Art}
We conclude by comparing our RAMP-VO with state-of-the-art methods on the Stereo DAVIS \cite{Zhou18eccv} dataset, the EDS \cite{eds} benchmark, and the Tartan Air test-split from the ECCV 2020 SLAM competition. We generate events for the ECCV2020 competition using VID2E \cite{Gehrig20cvpr} with default settings.
We run RAMP net asynchronously for all tests, using $M=20000$ for Stereo DAVIS \cite{Zhou18eccv} and $M=250000$ events with the ECCV2020 competition test-split and EDS. Poses are collected after processing a frame.

\input{tables/tab_EDS}

\textbf{Results:} 
Results for ECCV 2020 competition are available in Table \ref{tab:ECCVcompetition}, where we follow the evaluation in \cite{dpvo}, reporting the ATE[m] of the median of 5 runs with scale alignment. RAMP-VO is able to recover a better pose compared to all other image-based state-of-art methods \cite{orbslam3,dso,Schonberger_2016_CVPR,droidslam,dpvo} in most cases, outperforming by $19\%$ and $48\%$ DPVO \cite{dpvo} and DROID-SLAM using loop closure \cite{droidslam}, respectively.

Results on Stereo DAVIS are reported in Table \ref{tab:stereo_davis}. Given our emphasis on space applications, where space-graded cameras such as the AURICAM\textsuperscript{TM} \footnote{AURICAM's datasheet available on \href{https://sodern.com/wp-content/uploads/2023/11/2023-10-04-AURICAM-datasheet.pdf}{www.sodern.com}.}
often operate at low-FPS, we evaluate performance using the 5 FPS benchmark introduced in the supplementary analysis of \cite{eds}. Except for \emph{Bin}, the proposed RAMP-VO outperforms all other baselines that use either both or just one of the two modalities on Stereo DAVIS, surpassing the top-performing method, EDS, by $27.6\%$. On the EDS benchmark reported in Table \ref{tab:EDS}, our method consistently improves over DPVO and achieves an average 30\% improvement over DEVO which, contrary to our method, is trained on additional data from the TartanAir test set. These benchmarks represent completely new scenarios compared to the TartanAir training setting. Stereo DAVIS features gray-scale frames and a lower $180 \times 240$ resolution, while EDS has a higher resolution from Prophesee Gen3.1 and FLIR cameras, and an increased variety of test cases, including light and dark scenes and wider motions. Similar to the Apollo landing dataset, RAMP-VO is thus required to generalize from simulated to real sensor data.

It is worth noting how effective processing and fusion of event data is particularly important to achieve high performance in these benchmarks. Indeed, naive adaptations of DPVO for event processing fall short, while specialized event fusion techniques, like ours, can achieve better generalization and transfer to real-world data.

%% file: tables/tab_malapert_apollo.tex
\begin{table}
    \begin{center}
    \setlength{\tabcolsep}{3pt}
    \caption{Average Absolute Trajectory Error on the Malapert and Apollo datasets}
    \label{tab:malapert_apollo}
    \begin{tabular}{cc|cc|ccc} 
        & \multirow{2}{*}{Input} & \multicolumn{2}{c|}{Malapert [km]} & \multicolumn{3}{c}{Apollo [cm]} \\
        &                        & cam-1 & cam-2 &  rec-1 &  rec-3 &  rec-4\\ 
        \hline
        DPVO \cite{dpvo}                     & I    & 73.1 &  48.2   & 0.9  &  0.3 & 1.3\\ 
        DPVO \cite{dpvo}                    & I+E  & 40.1 &  34.0 & 0.9  &  0.2 & 1.1 \\ 
       \textbf{RAMP-VO SS (Ours)}           & I+E  & 1.1  &  9.4 & \textbf{0.7} & 0.2 & 1.0\\ 
        \textbf{RAMP-VO (Ours)}              & I+E  &\textbf{0.6} & \textbf{4.3} & 0.8 & \textbf{0.2} & \textbf{0.9} \\ \hline
    \end{tabular}
    \end{center}
    \vspace{-1em}
\end{table}

%% file: tables/tab_tartanAIR_competition.tex
\renewcommand*\rot[2]{\multicolumn{1}{R{#1}{#2}}}%
\newcommand\rotmulti[1]{\rotatebox{90}{\parbox{1.6cm}{\centering#1}}}

\begin{table}%
    \begin{center}
    \caption{Average Absolute Trajectory Error (m) on ECCV 2020 SLAM competition monocular test-split. Methods marked with (\cmark) use global optimization / loop closure. Top performing (non-global) method in bold, second best underlined.} 
    \label{tab:ECCVcompetition}    

    \begin{tabular}{c|ccc|cccc}
\textbf{} & \textbf{\rotmulti{ORB-SLAM3 \cite{orbslam3}}} & \textbf{\rotmulti{COLMAP \cite{Schonberger_2016_CVPR}}} & \textbf{\rotmulti{DROID-SLAM \cite{droidslam}}} & \textbf{\rotmulti{DSO \cite{dso}}} & \textbf{\rotmulti{DROID-VO \cite{droidslam}}} & \textbf{\rotmulti{DPVO \cite{dpvo}}} & \textbf{\rotmulti{RAMP-VO (Ours)}} \\ \hline
Input     & I                                                 & I                                                            & I                                                    & I                                      & I                                & I                                        & I+E                                             \\
Global    & \cmark                                            & \cmark                                                       & \cmark                                               & \xmark                                 & \xmark                           & \xmark                                   & \xmark                                          \\ \hline
ME00      & 13.61                                             & 15.20                                                        & 0.17                                                 & 9.65                                   & 0.22                             & \textbf{0.16}                            & \underline{0.20}                                            \\
ME01      & 16.86                                             & 5.58                                                         & 0.06                                                 & 3.84                                   & 0.15                             & \underline{0.11}                         & \textbf{0.04}                                   \\
ME02      & 20.57                                             & 10.86                                                        & 0.36                                                 & 12.20                                  & 0.24                             & \underline{0.11}                         & \textbf{0.10}                                   \\
ME03      & 16.00                                             & 3.93                                                         & 0.87                                                 & 8.17                                   & 1.27                             & \underline{0.66}                         & \textbf{0.46}                                   \\
ME04      & 22.27                                             & 2.62                                                         & 1.14                                                 & 9.27                                   & 1.04                             & \underline{0.31}                         & \textbf{0.16}                                   \\
ME05      & 9.28                                              & 14.78                                                        & 0.13                                                 & 2.94                                   & \underline{0.14}                 & \underline{0.14}                         & \textbf{0.13}                                   \\
ME06      & 21.61                                             & 7.00                                                         & 1.13                                                 & 8.15                                   & 1.32                             & \underline{0.30}                         & \textbf{0.12}                                   \\
ME07      & 7.74                                              & 18.47                                                        & 0.06                                                 & 5.43                                   & 0.77                             & \underline{0.13}                         & \textbf{0.12}                                   \\ \hline
MH00      & 15.44                                             & 12.26                                                        & 0.08                                                 & 9.92                                   & \underline{0.32}                 & \textbf{0.21}                            & 0.36                                            \\
MH01      & 2.92                                              & 13.45                                                        & 0.05                                                 & 0.35                                   & 0.13                             & \textbf{0.04}                            & \underline{0.06}                                \\
MH02      & 13.51                                             & 13.45                                                        & 0.04                                                 & 7.96                                   & \underline{0.08}                 & \textbf{0.04}                            & \textbf{0.04}                                   \\
MH03      & 8.18                                              & 20.95                                                        & 0.02                                                 & 3.46                                   & 0.09                             & \underline{0.08}                         & \textbf{0.04}                                   \\
MH04      & 2.59                                              & 24.97                                                        & 0.01                                                 & -                                      & 1.52                             & \underline{0.58}                         & \textbf{0.41}                                   \\
MH05      & 21.91                                             & 16.79                                                        & 0.68                                                 & 12.58                                  & 0.69                             & \textbf{0.17}                            & \underline{0.25}                                \\
MH06      & 11.70                                             & 7.01                                                         & 0.30                                                 & 8.42                                   & \underline{0.39}                 & \textbf{0.11}                            & \textbf{0.11}                                   \\
MH07      & 25.88                                             & 7.97                                                         & 0.07                                                 & 7.50                                   & 0.97                             & \underline{0.15}                         & \textbf{0.07}                                   \\ \hline
Average   & 14.38                                             & 12.50                                                        & 0.33                                                 & 7.32                                   & 0.58                             & \underline{0.21}                         & \textbf{0.17} \\ \hline                                 
\end{tabular}%
\end{center}
\end{table}

%% file: tables/tab_stereo_davis.tex
\begin{table}
    \begin{center}
    \caption{Average Absolute Trajectory Error (cm) on Stereo Davis } \label{tab:stereo_davis}
    \begin{tabular}{cc|cccc} 
                   & Input & Bin & Boxes & Desk &  Monitor\\  \hline
        ORB-SLAM2 \cite{MurArtal17tro}
                   & I     & \textbf{2.5} &  7.0  & 9.3  &  10.3\\ 
        DPVO \cite{dpvo}      
                   & I     & 3.6 &  6.7  & 7.0  &  11.5\\ \hline
        DPVO \cite{dpvo}
                   & E     & 5.8 & 7.2   & 7.8  &  14.4\\ \hline
        EDS \cite{eds}
                   & I+E   & 2.6 & 5.8 & 5.0 & 8.0 \\ 
        DPVO \cite{dpvo}
                   & I+E   & 4.2 & 5.3 & 4.9 & 6.2 \\ 
        \textbf{RAMP-VO (Ours)}
                   & I+E   & 3.1 &  \textbf{5.1} & \textbf{3.1} & \textbf{4.2} \\ \hline
    \end{tabular}
    \end{center}
\end{table}

%% file: tables/tab_EDS.tex
\begin{table}%
    \begin{center}
    \vspace{-3pt}
    \caption{Avg. Absolute Trajectory Error (cm) and $R_{rmse}$ (deg) on EDS. ($^*$) indicates methods trained on TartanAir train+test data. Top performing (non-global) VO method in bold, second best underlined.} 
    \label{tab:EDS}    
    
    \setlength{\tabcolsep}{3.5pt} %
    \begin{tabular}{c|c|ccc}
    
\textbf{} & \textbf{\rotmulti{ORB-SLAM3 \cite{orbslam3}}} & 
\textbf{\rotmulti{DPVO \cite{dpvo}}} & 
\textbf{\rotmulti{DEVO$^*$ \cite{klenk2023deep}}} & 
\textbf{\rotmulti{RAMP-VO (Ours)}} 
\\ \hline
Input     & I & I & E & I+E  \\
Global    & \cmark & \xmark & \xmark  & \xmark \\ \hline
 & ATE / $R_{rmse}$ & ATE / $R_{rmse}$ & ATE / $R_{rmse}$ & ATE / $R_{rmse}$ \\ \hline
pean. dark    & 6.15 / 11.40        & \underline{1.26} / \underline{1.83}   & 4.78 / 2.49                         & \textbf{1.20} / \textbf{0.64} \\ 
pean. light   & 27.26 / 6.88        & \underline{12.99} / \textbf{2.66}     & 21.07 / \underline{3.84}            & \textbf{9.03} / 6.40 \\ 
pean. run     & 16.83 / 5.78        & \underline{25.48} / \textbf{11.19}    & 38.10 / 18.28                       & \textbf{13.19} / \underline{11.43} \\ 
rocket dark   & 10.12 / 9.75        & 27.41 / 5.23                          & \underline{8.78} / \underline{4.16} & \textbf{7.20} / \textbf{2.63} \\ 
rocket light  & 32.53 / 11.39       & 63.11 / 10.44                         & \underline{59.83} / \underline{9.28}& \textbf{17.53} / \textbf{4.04} \\ 
ziggy         & 26.92 / 4.42        & \underline{14.86} / \underline{3.45}  & \textbf{11.84} / \textbf{2.32}      & 19.05 / 7.66 \\ 
ziggy hdr     & 81.98 / 17.67       & 66.17 / 10.32                         & \textbf{22.82} / \underline{9.07}   & \underline{28.78} / \textbf{5.13} \\ 
ziggy fly.    & 20.57 / 8.02        & \underline{10.85} / \underline{3.66}  & 10.92 / \textbf{3.39}               & \textbf{6.35} / 5.07 \\ 
all chars     & 21.37 / 9.02        & 95.87 / 29.00                         & \textbf{10.76} / \textbf{3.62}      & \underline{28.61} / \underline{9.89} \\ \hline
Average       & 27.08 / 9.37        & 35.33 / 8.64                          & \underline{21.00} / \underline{6.27}& \textbf{14.57} / \textbf{5.88}\\ \hline            
\end{tabular}
\end{center}
\end{table}

%% file: sections/5.conclusions.tex
\section{Conclusion}
\label{sec:conclusions}

In this work, we introduce RAMP-VO, an end-to-end VO system tailored for challenging environments such as those encountered during lunar descents. 
RAMP-VO uses RAMP encoders to fuse event data into frames, achieving $8\times$ speedup and 33\% improvement over state-of-the-art asynchronous encoders. 
Moreover, by incorporating events, RAMP-VO reduces the trajectory error of existing image-only deep-learning-based solutions by 58.8\%, as well as event-only methods by 30.6\% on average. Experiments show that RAMP-VO can transfer zero-shot to real data despite being trained only on synthetic one, while still outperforming the other baselines. 
We designed RAMP-VO %
for space applications focusing on improving accuracy and sensor fusion. Future efforts should prioritize optimizing the method for deployment on resource-constrained, space-graded hardware leveraging techniques such as model quantization, network compression, and efficient hardware implementations \cite{furano2020towards}.
Despite these limitations, we view this work as a milestone in event data fusion for VO, and we believe it can spark new interest in the use of event cameras and learning-based approaches for robust navigation.